\documentclass[preprint,12pt]{elsarticle}
\usepackage{amsmath, bm}
\usepackage{hyperref}

\usepackage{amssymb}
\usepackage{appendix}
\usepackage{lineno}
\setcitestyle{authoryear,open={(},close={)}}

\journal{Environmental Modelling \& Software}

\begin{document}

\begin{frontmatter}



\title{Long-term drought prediction using deep neural networks based on geospatial weather data}


\author[inst1]{Alexander Marusov \corref{contrib}}

\author[inst1]{Vsevolod Grabar \corref{contrib}}

\author[inst2]{Yury Maximov}
%\ead{yury@lanl.gov}

\author[inst3]{Nazar Sotiriadi}
%\ead{n.sotiriadi@gmail.com}

\author[inst1]{Alexander Bulkin}
%\ead{A.Bulkin@skoltech.ru}

\author[inst1]{Alexey Zaytsev}
%\ead{a.zaytsev@skoltech.ru}

% \cortext[contrib]{\phantom{$aaaaaaaaaaaaaa$} \\ *Authors contributed equally}

\cortext[contrib]{Authors contributed equally}

% \fntext[label1]{Methodology}
% \fntext[label2]{Software}
% \fntext[label3]{Validation}
% \fntext[label4]{Visualization}
% \fntext[label5]{Writing - Original Draft}
% \fntext[label6]{Writing - Review \& Editing}
% \fntext[label7]{Conceptualization}
% \fntext[label8]{Conceptualization}

\affiliation[inst1]{organization={Skolkovo Institute of Science and Technology}}

\affiliation[inst2]{organization={Los Alamos National Laboratory, Theoretical Division},
    %city={Los Alamos},
    %postcode={NM 87545}, 
    %country={USA}}
}
\affiliation[inst3]{organization={Sberbank of Russia PJSC},
    %city={Moscow},
    %postcode={117312}, 
    %country={Russian Federation}}
}
%\affiliation[inst4]{organization={Moscow State University},
    % addressline={Staromonetniy pereulok 29}, 
    %city={Moscow},
    %postcode={119234}, 
    %country={Russian Federation}}
%}
% \affiliation[inst2]{organization={Skolkovo Institute of Science and Technology}}
% \affiliation[inst3]{organization={Skolkovo Institute of Science and Technology}}

% \author[inst2]{Author Two}
% \author[inst1,inst2]{Author Three}

% \affiliation[inst2]{organization={Department Two},%Department and Organization
%             addressline={Address Two}, 
%             city={City Two},
%             postcode={22222}, 
%             state={State Two},
%             country={Country Two}}

\begin{abstract}
The problem of high-quality drought forecasting up to a year in advance is critical for agriculture planning and insurance. Yet, it is still unsolved with reasonable accuracy due to data complexity and aridity stochasticity. We tackle drought data by introducing an end-to-end approach that adopts a spatio-temporal neural network model with accessible open monthly climate data as the input.

Our systematic research employs diverse proposed models and five distinct environmental regions as a testbed to evaluate the efficacy of the Palmer Drought Severity Index (PDSI) prediction. Key aggregated findings are the exceptional performance of a Transformer model, EarthFormer, in making accurate short-term (up to six months) forecasts. At the same time, the Convolutional LSTM excels in longer-term forecasting. 
\end{abstract}


% \textbf{Highlights}:

% \begin{enumerate}

%     \item We adopted spatiotemporal methods (ConvLSTM, gradient boosting, logistic regression).

%     \item We made work modern transformers (EarthFormer, FourCastNet).

%     \item We reduced the gap to perfect ROC-AUC by $54\%$ and $16\%$, respectively.

%     \item We created an extensive test bed to evaluate and compare our models objectively.
% \end{enumerate}


\end{frontmatter}




\section{Highlights}

\begin{itemize}

\item We improved quality for long-term, up to 12 months drought forecasting

\item We adopted modern transformers and Convolutional LSTM to solve this problem

\item We created an extensive test bed to evaluate models consisting of 5 diverse regions

\item We reduced the gap to perfect ROC-AUC by $54\%$ and $16\%$, respectively

\end{itemize}

\section{Software and data availability}

\begin{itemize}
    \item Software name: Long-term drought prediction

    \item Developer: Vsevolod Grabar [repo creator, contribution], Alexander Marusov [contribution]

    \item Contact information: \href{mailto:astralex98@gmail.com}{astralex98@gmail.com}

    \item First year available: 2023

    \item Program language: Python

    \item Cost: free

    \item Software and data availability: \footnote{\url https://github.com/Astralex98/long-term-drought-prediction/tree/main}

    \item Repository storage: $120$ MB
    
\end{itemize}

\section{CRediT author statement}
\textbf{Alexander Marusov:} Methodology, Software, Validation, Visualization, Writing – original draft, Writing – review \& editing. \textbf{Vsevolod Grabar:} Software, Validation, Visualization, Writing – original draft, Writing – review \& editing. \textbf{Yury Maximov:} Conceptualization. \textbf{Nazar Sotiriadi:} Conceptualization. \textbf{Alexander Bulkin:} Conceptualization. \textbf{Alexey Zaytsev:} Conceptualization, Funding acquisition, Methodology, Project administration, Resources, Supervision, Validation, Writing - Original Draft, Writing - Review \& Editing.

\section{Introduction}
\label{sec:intro}

The forecasting of droughts represents a critical challenge in climate science~\citep{safwan2022compare}, as these natural phenomena incur substantial losses and can significantly impact populations and various economic sectors \citep{adikari2021ai}. 
The importance of monitoring and predicting droughts is underscored by their frequent occurrence in diverse geographical landscapes \citep{ghozat2023drought}. 
Moreover, the likelihood of droughts is expected to increase in the context of global climate change \citep{Xiujia2022warming}. 
Their accurate forecasting, however, is a complex problem due to the inherent difficulty in predicting the onset, duration, and cessation of drought events \citep{mishra2005stochastic}. Another difficulty lies in choosing a drought index suitable for the goal being targeted.

% What is PDSI, and why do we need it
We focus on long-term decision-making, which is critical for the annual planning of agricultural and insurance companies \citep{zhang2019climate}. 
Formally, it is desired to provide accurate forecasts of droughts that extend 12 months into the future. 
A particular drought severity index for prediction is also important, as various indexes take into account diverse climatic factors, including temperature and precipitation. 
Among the drought severity indices, the Standardized Precipitation Index (SPI) \citep{mckee1993spi} and the Palmer Drought Severity Index (PDSI) \citep{alley1984palmer} stand out as fundamental measures. Another example of a modern drought index is the Standardized Precipitation-Evapotranspiration Index (SPEI) \citep{vic2010spei}. 

Our research utilizes the monthly PDSI for several compelling reasons. 
Firstly, given our extended forecast horizon of 12 months, PDSI is recognized as an effective tool for long-term assessment \cite{mcpherson2022pdsiforecast}. 
Additionally, PDSI's extensive historical record facilitates an analysis of the impacts of global warming within the broader climate dynamics \cite{dai2011pdsi}. 
It is known that drought substantially can be divided into several types: meteorological, agricultural, hydrological, and socioeconomic ~\citep{hao2017drought}. 
As our study concentrates on meteorological and agricultural droughts, leaving hydrological and socioeconomic issues out of scope, PDSI emerges as the most pertinent index, aligning well with the types of droughts we are investigating.
Finally, we aim for a more interpretable forecast and quantifying PDSI values into the selected bins corresponding to different drought severity levels. So, instead of predicting PDSI directly (a regression task), we treat our task as a classification problem predicting bins and estimating the probability of severe drought, which is the most interpretable quantity for decision-makers.

% What models are used for drought prediction
% GCM methods
There are numerous approaches already available to solve the problem at hand.
As a natural climate phenomenon, drought could be evaluated with the help of various climate models.
General Circulation Models (GCMs) are powerful modern methods for climate event prediction, utilizing partial differential equations to simulate Earth's systems \citep{jiang2021wavelet}. GCMs typically model the Earth's climate with a three-dimensional grid of spatial resolution between 250 and 600 km, 10 to 20 layers in the atmosphere, and up to 30 layers in the oceans. \citet{wang2014cmip5} used 35 GCMs from Coupled Model Intercomparison Project Phase 5 (CMIP5) to estimate increasing drought frequency in China. Similarly, \citet{song2022differences} compared SPI and SPEI drought indices in South Korea using CMIP5 and CMIP6 GCMs, revealing varied forecast reliability over different time frames. 
While GCMs offer comprehensive global climate insights over decades-long horizons, their inherent uncertainties, especially in predicting extreme events, and lower resolution limit their effectiveness for regional climate analysis.

% ML approaches
An alternative is machine learning algorithms for drought forecasting tasks \citep{prodhan2022review}. 
Classical approaches, such as stochastic models like AutoRegressive Integrated Moving Average (ARIMA) and multiplicative Seasonal AutoRegressive Integrated Moving Average (SARIMA), have been effectively utilized for drought prediction using the SPI index \cite{mishra2005stochastic}. 
In another instance, multivariate regression, incorporating historical PDSI data and other global climate indices, was employed to forecast the PDSI index in South-Central Oklahoma \citep{mcpherson2022pdsiforecast}. While these methods predominantly address regression problems, we focus on classification.
Logistic regression has been applied for binary drought classification using the SPI index in \cite{niaz2021logreg}, demonstrating its suitability for label prediction tasks. 
Also, the gradient boosting algorithm has emerged as a powerful tool in modeling geospatial data \cite{proskura2019usage,koldasbayeva2022large}, especially effective in handling the imbalanced datasets often encountered in drought prediction \citep{kozlovskaia2017deep}. Notably, this algorithm has also remarkably succeeded in classifying drought conditions in Turkish regions \citep{mehr2021xgboost}.

% Deep learning methods for drought forecasting
Addressing the needs of practitioners, deep learning emerged as a viable tool for drought forecasting. \citet{mishra2006feedforward} introduced the use of recursive and direct multistep neural networks, leveraging the SPI index. Among time-series data approaches, Recurrent Neural Networks (RNN) \citep{Rumelhart1985} stand out, frequently outperforming traditional methods in time-series analysis \citep{Hewamalage2021}. Specifically, Long Short-Term Memory (LSTM) networks \citep{hochreiter1997lstm} have been shown to surpass ARIMA in long-term SPI index forecasting, although ARIMA remains competitive in short-term prediction \citep{poornima2019lstm}. To harness the strengths of both ARIMA and LSTM, \citet{xu2022arimalstm} proposed a hybrid ARIMA-LSTM model.

% Spatio-temporal modeling
However, these methods primarily focus on historical (temporal) data, neglecting the spatial aspects. Addressing both temporal and spatial dependencies, the ConvLSTM method has been applied to various fields, including precipitation prediction~\citep{shi2015convolutional}, earthquake prediction \citep{kail2021recurrent}, and was notably used by \cite{park2020convlstm} for short-term (eight-day) drought forecasting using satellite-based indices like Scaled Drought Condition Index (SDCI) and SPI.
The emergence of Transformer architectures, originally developed for Natural Language Processing (NLP) \citep{vaswani2017attention}, led to new models in diverse domains such as Computer Vision (CV) \citep{zheng2021poseformer}. This makes their adoption of spatiotemporal modeling a compelling choice. Prominent examples include EarthFormer \citep{gao2022EarthFormer} and FourCastNet \citep{pathak2022fourcastnet}, which excel in various spatiotemporal tasks, including regression challenges like precipitation nowcasting, while these approaches can suffer from low amount of available training data and high stochasticity of a target in a long-term forecasting problem. In our research, we have adapted these advanced Transformer architectures to enhance drought forecasting, aiming to leverage their capabilities in handling complex spatiotemporal data.

In the paper, we build spatio-temporal models for PDSI index forecasting, solving the long-term drought prediction problem and obtaining the following key contributions:
\begin{enumerate}
    
    \item Comprehensive study of diverse spatio-temporal models. We adapted advanced deep-learning methods from different domains, benchmarking the most prominent options, including transformer-based models EarthFormer and FourCastNet. Also, classic approaches (logistic regression and gradient boosting) were constructed to account for both spatial and temporal dependencies. According to our knowledge, we are the first to adopt the above-mentioned spatio-temporal neural network models for long-term drought forecasting within the PDSI index with notable quality of the developed models.
    \item Development of a test bed encompassing five distinct global regions for objective model evaluation with publicly accessible data. Together with a wide range of compared models, this research provides a systematic answer to the question of how one should predict droughts one to 12 months ahead and do we need deep learning to provide accurate forecasts.
    \item Identification of our neural networks based on EarthFormer as optimal for medium-term forecasting up to six months and our variant ConvLSTM for long-term predictions. Thus, we make transformers work for a relatively small amount of input and training data. The input data for our models are easy to obtain and preprocess, making the model straightforward to run and more robust compared to elaborated preprocessing and feature engineering used in previous studies.
    While logistic regression and gradient boosting are enough in short-term forecasting, deep learning methods excel in four-month and longer time frames, which is crucial for decision-making.
    \item Consistent improvement of models for horizons ranging from 1 to 12 months. This focus contrasts with previous studies, which typically addressed either very short-term predictions (up to a month) or much longer-term forecasts spanning decades.
\end{enumerate}

% The following are the key \textbf{highlights} of our work:

% \begin{enumerate}

%     \item We adopted spatiotemporal methods (ConvLSTM, gradient boosting, logistic regression).

%     \item We made work modern transformers (EarthFormer, FourCastNet).

%     \item We reduced the gap to perfect ROC-AUC by $54\%$ and $16\%$, respectively.

%     \item We created an extensive test bed to evaluate and compare our models objectively.
% \end{enumerate}

\section{Data}
 In this section, we consider the definition of the selected target variable PDSI, drought classification based on its values, and the characteristics of input features for model prediction.

The Palmer Drought Severity Index (PDSI) is a standardized index where absolute values above 4 indicate extreme conditions, and intermediate numbers are divided into bins and assigned to various wet or dry environments, where the latter corresponds to negative PDSI, see Table \ref{tab:pdsi_bins}. It is calculated using a version of the Palmer formula, which combines reference evapotranspiration, precipitation, and a static soil water-holding capacity layer.

\begin{table}[!h]
\small
    \centering
    \begin{tabular}{l@{\hskip 1.0in}l}
         \hline
         PDSI value & Drought severity class \\
         \hline 
         4.00 and above & Extreme wet spell    \\
         3.00-3.99 & Severe wet spell   \\
         2.00-2.99 & Moderate wet spell \\
         1.00-1.99 & Mild wet spell  \\
         -1.00 to 0.99 & Normal  \\
         -1.00 to -1.99 & Mild dry spell  \\
         -2.00 to -2.99 & Moderate dry spell  \\
         -3.00 to -3.99 & Severe dry spell  \\
         -4.00 and below & Extreme dry spell  \\
         \hline
    \end{tabular}
    \caption{Classification of various PDSI values, source:~\cite{liu2015}}
    \label{tab:pdsi_bins}
\end{table}

\begin{figure}[h]
    \centering
    \includegraphics[width=\textwidth]{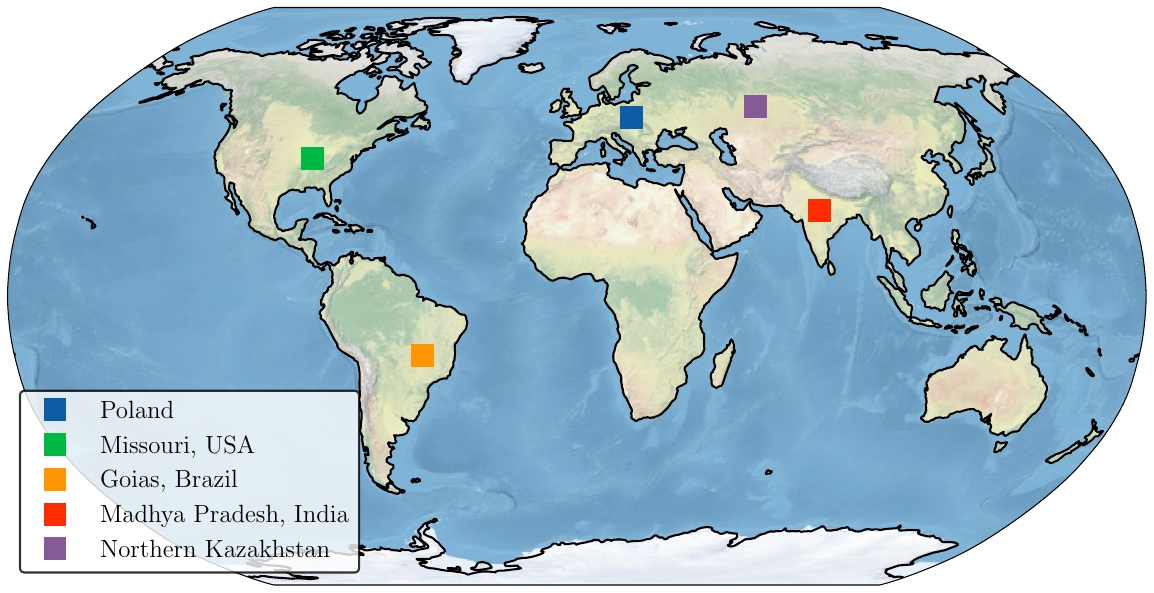}
    \caption{Diverse regions chosen for PDSI forecast}
    \label{fig:worldmap}
\end{figure}

We have used publicly available geospatial data from Google Earth Engine~\citep{gorelick2017google}. 
Specifically, to obtain the PDSI data, we employed the TerraClimate Monthly dataset~\footnote{\url{https://developers.google.com/earth-engine/datasets/catalog/IDAHO_EPSCOR_TERRACLIMATE}}. 
Our PDSI data encompasses a comprehensive range of climatic values from 1958 to 2022, covering the Earth's entirety. 
To test the consistency of the considered models, we examined regions across continents and climate zones: from the state of Missouri to Poland to India.
The considered regions are depicted in Figure~\ref{fig:worldmap}, and their characteristics are shown in Table~\ref{tab:summary_stats}.

The input data, downloaded in a tif format, were transformed into a 3D tensor containing PDSI values. This tensor's structure includes one time-scale dimension (monthly intervals) and two spatial dimensions (x and y coordinates) representing the grid. 
The regional datasets exhibit varied resolutions, with grid dimensions ranging from 30 x 60 to 128 x 192 cells, allowing for varying degrees of granularity and spatial detailing. 
The spatial resolution for a cell in the TerraClimate data is approximately 5 km.

\begin{table}[!h]
\small
    \centering
    \begin{tabular}{lcccc}
    \hline
    Region & Span, & $\%$ of normal & $\%$ of drought & Spatial sizes \\
    & months & PDSI $\geq -2$  & PDSI $\leq -2$ & km$\times$km \\
    \hline
    Missouri, USA & 754 & 74.91 & 25.09 & $416\times544$\\
    Madhya Pradesh, India & 754 & 70.66 & 29.34 & $512\times768$\\
    Goias, Brazil & 754 & 68.97 & 31.03 & $640\times640$\\
    Northern Kazakhstan & 742 & 68.70 & 31.30 & $256\times480$\\
    Poland & 742 & 66.28 & 33.72 & $352\times672$ \\
    \hline
    \end{tabular}
    \caption{Regions' summary statistics}
    \label{tab:summary_stats}
\end{table}

\section{Methods}

We compare deep learning approaches, including Convolutional LSTM and novel transformer models, such as FourCastNet from Nvidia and EarthFormer from Amazon, with classic methods, including the baseline model, gradient boosting, and logistic regression.

\subsection{Baseline}
As a global baseline and a coherence check, we took the most prevalent class from the training data as the prediction and compared it with actual targets from the test subset. We also checked a rolling baseline  (i.e., the most frequent class from recent history, from 6 to 24 months). Still, the results were almost indistinguishable from the global baseline, so we did not include them in our tables and graphs.

\subsection{Basic methods: Logistic regression and Gradient boosting}
Both logistic regression and gradient boosting cannot work with raw data. Therefore, we created a data module that treated each grid cell as an individual value and transformed our task into a typical time series forecasting problem. To benefit from spatial correlations, we incorporate values from neighboring cells. For example, if we consider a 3x3 cell neighborhood, this includes eight additional time series. It is important to note that for "edge" cells, some of the neighboring cells may contain all zeros due to data limitations.

\paragraph{Logistic Regression}
Logistic regression is usually the natural choice for tasks with linear dependence or as a baseline model. The novel research ~\cite{zeng2023linear} shows that time series linear models are a good choice.

\paragraph{Gradient boosting}
We adopted the gradient boosting of decision trees, implemented using the well-established library XGBoost~\cite{chen2016xgboost}. 
XGBoost is renowned for its speed and efficiency across a wide range of predictive modeling tasks and has consistently been favored by data science competition winners. It operates as an ensemble of decision trees, with new trees aimed at rectifying errors made by existing trees in the model. Trees are successively added until no further improvements can be made. 

\subsection{Convolutional LSTM (ConvLSTM)}
\begin{figure}[h]
    \centering
    \includegraphics[scale=0.40]{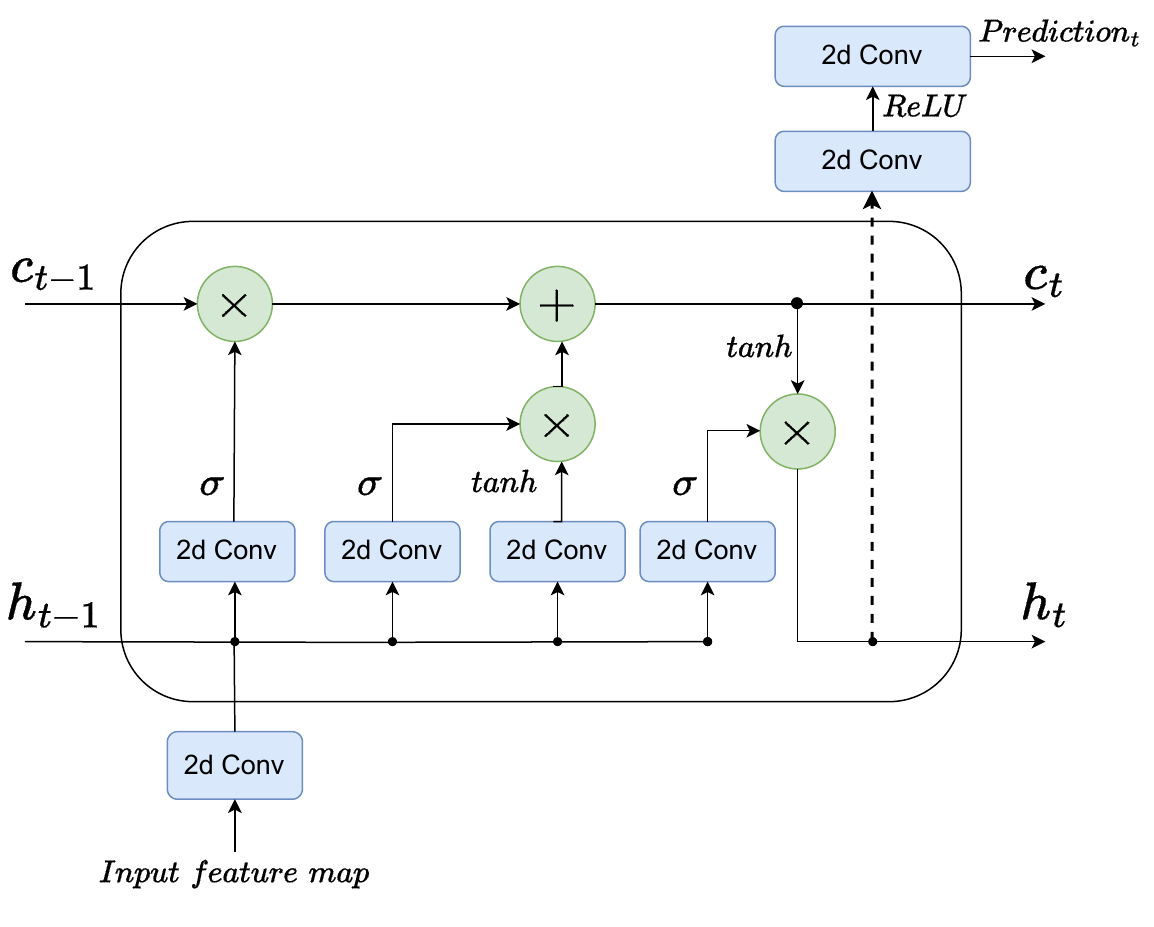}
    \caption{Our version of ConvLSTM architecture}
    \label{fig:convlstm}
\end{figure}

Our model, inspired by \cite{kail2021recurrent}, modifies the Convolutional LSTM architecture (\cite{shi2015convolutional}), blending Recurrent Neural Networks (RNNs) with Long Short-Term Memory (LSTM) to capture temporal dependencies. This adaptation involves extending LSTM's traditional one-dimensional hidden states to two-dimensional feature maps, facilitating grid-to-grid transformations essential for spatial tasks. We process PDSI grids varying from $50 \times 50$ to $200 \times 200$ cells, integrating Convolutional Neural Networks (CNNs) for spatial analysis. This combination, depicted in Figure~\ref{fig:convlstm}, leverages both RNN and CNN strengths, simultaneously capturing temporal and spatial information in the architecture for drought prediction.

\paragraph{Details of architecture}

Convolutional LSTM follows the pipeline:
\begin{enumerate}
    \item  Represent data as a sequence of grids: For each cell, we specify a PDSI value for a particular month; the input grid at each time moment has dimension $grid_h \times grid_w$ (varying from $50 \times 50$ to $200 \times 200$ for different regions of interest).
    \item Pass the input grid through a convolutional network to create an embedding of grid dimensionality with 16 channels. As an output of LSTM at each time moment, we have a hidden representation (short-term memory) of size $hidden \times grid_h \times grid_w$, cell (long-term memory) representation of a similar size, and the output of size $hidden\times grid_h \times grid_w$. 
    \item Transform the output to the size $1 \times grid_h \times grid_w$ using convolution $1 \times 1$ to receive probabilities of the drought for each cell as a final prediction or to $k\times grid_h \times grid_w$ in case of multiclass classification, where $k > 2$ is the number of classes of drought condition that we are trying to predict. As an additional hyperparameter, we vary the forecasting horizon (to forecast PDSI for the next month or $f$-th month).
\end{enumerate}

\subsection{Transformer-based methods}

We consider two recent transformer-based models, FourCastNet and EarthFormer, adopted for usage in spatiotemporal problems such as drought prediction.

\paragraph{FourCastNet}
 Fourier ForeCasting Neural Network (FourCastNet) was developed by ~\cite{pathak2022fourcastnet} as a weather forecasting model and a part of NVIDIA's Modulus Sym deep learning framework for solving applied physics tasks. This model combines Adaptive Fourier Neural Operator (AFNO) from~\cite{guibas2022adaptive} with Vision Transformer. The model is computationally and memory efficient, with low spatial mixing complexity of $O(N\log N)$, where $N$ is the sequence length. The authors produced high-resolution short-term wind-speed and precipitation forecasts in the original papers. We modified the last layer of the model, switching it from a regression to a classification task and evaluating it for the long-term forecasting problem with a significantly smaller amount of available data.

\paragraph{EarthFormer}
Vanilla transformers have $O(N^2)$ complexity, and consequently, it is hard to apply them to spatiotemporal weather data because of their large dimensionality. \cite{gao2022EarthFormer} suggest using the "divide and conquer" method: they divide the original data into non-intersecting parts (called cuboids) and use a self-attention mechanism on each cuboid separately in parallel. Such an approach allows significantly reduced complexity, bridging the gap between transformers and CNNs. The authors introduced EarthFormer for regression tasks, and we adopted it for a classification problem in a way similar to FourCastNet. 

\subsection{Technical details}

Hyperparameters of deep learning models are mostly taken from corresponding original papers and can be found in our GitHub. This section in Table \ref{tab:optimization} presents some necessary optimization settings for all our deep learning methods.

\begin{table}[htb]
    \centering
    \resizebox{\textwidth}{!}{\begin{tabular}{lcccc}
    \hline
      & \multicolumn{4}{c}{Optimization}\\
     \hline
     Models & Epochs num & Batch & Optimizer & Learning rate scheduler \\
     & (average) & & & \\
     \hline
     ConvLSTM & 70 & 8 & Adam & absent \\
     EarthFormer & 20 & 16 & AdamW & Cosine \\
     FourCastNet & 30 & 8 & Adam & Cosine \\
     \hline
    \end{tabular}}
    \caption{Optimization characteristics}
    \label{tab:optimization}
\end{table}
\section{Results}

\paragraph{Formal problem statement}
We define a binary classification for drought forecasting using a PDSI threshold of $-2$, aligning with \cite{mcpherson2022pdsiforecast} and PDSI bin categorizations presented in Table~\ref{tab:pdsi_bins}. 
The model's objective is to predict drought occurrences, classified as serious, when PDSI falls below this threshold. 

For validation, we divided each local dataset (that corresponds to one of the five regions including Missouri, Northern Kazakhstan, Madhya Pradesh, Eastern Europe, and Goias) into training ($70\%$) and testing ($30\%$) subsets. The splitting uses out-of-time validation where the test set follows the training set chronologically. Model training and subsequent quality metric evaluations are conducted using these distinct data subsets.

\paragraph{Evaluation procedure}
We use $ROC\,AUC$, $PR\,AUC$ and $F1$ scores to evaluate the model.
During validation for early stopping and hyperparameter optimization, we chose the  $ROC\,AUC$ score. 
All these scores are the medians of all spatial predictions (because we want to remove the impact of outliers ), as we compute a temporal vector at every spatial prediction cell. Next, we receive a single score for each cell, so we end up with a grid of metrics. Finally, we get the median of scores at each set. 
Higher values for all scores correspond to better models.
$ROC\,AUC$ scores have the perfect value of $1$ and the value for a random prediction of $0.5$.

\paragraph{Compared methods}

For our main experiments, we explored the baseline's performance (the most frequent class from historical data), gradient boosting (XGBoost), logistic regression, and our modifications of ConvLSTM, FourCastNet, and EarthFormer.

XGBoost and logistic regression are two basic algorithms often used as strong baselines.
The last three are variants of neural networks that performed strongly in various geospatial problems.
They represent two dominating architectures in geospatial modeling: ConvLSTM is a combination of recurrent and convolutional neural networks; FourCastNet and EarthFormer are Transformers. 

\subsection{Main results}

\begin{figure}[!h]
    \centering
\includegraphics[width=.32\textwidth]{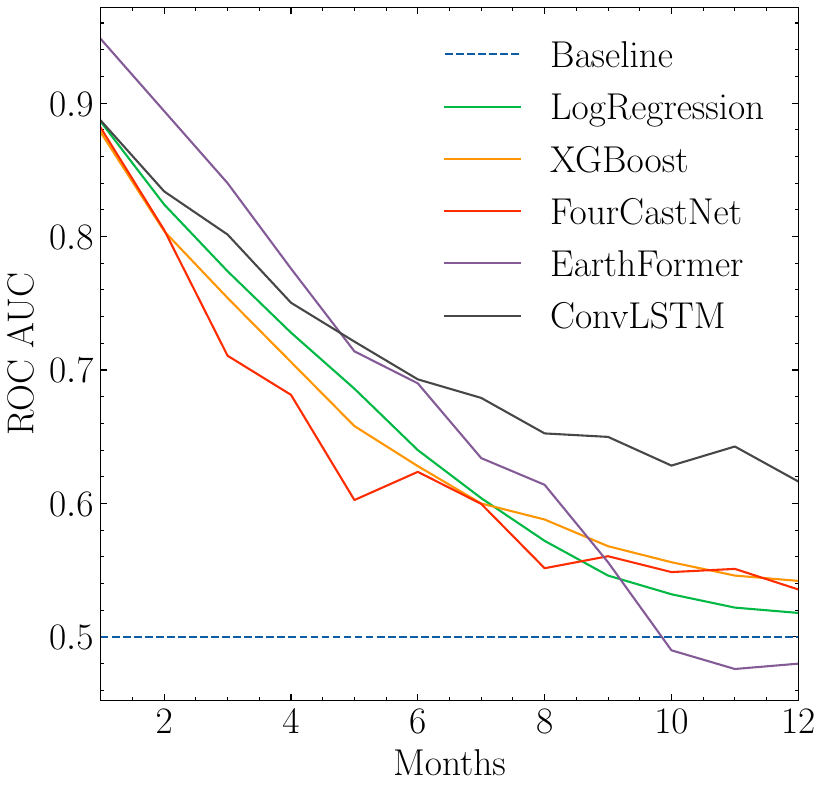}
\includegraphics[width=.32\textwidth]{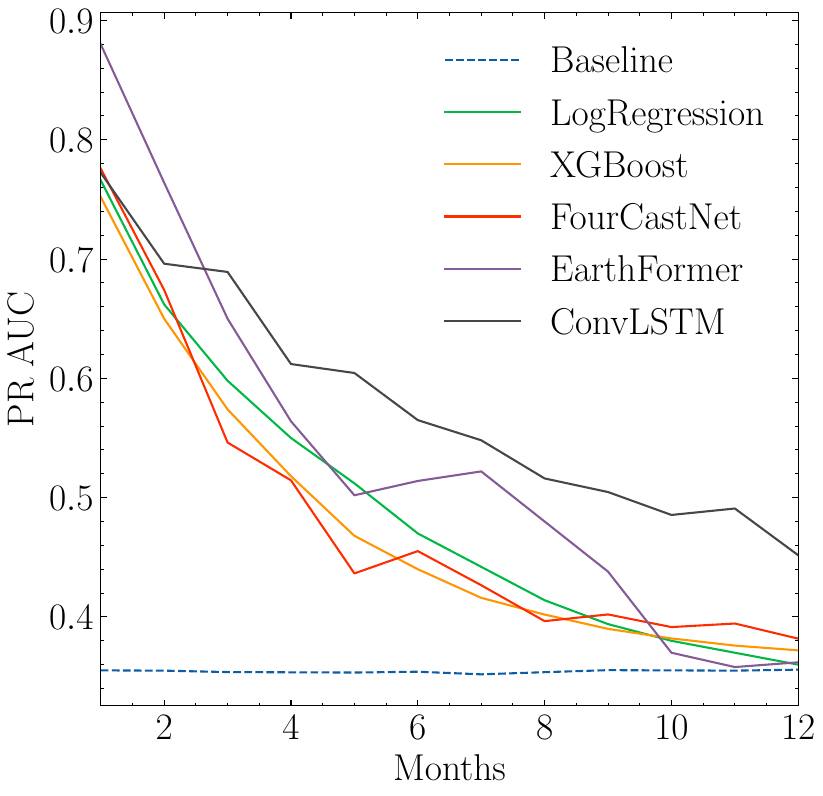}
%\\
%\includegraphics[width=.45\textwidth]{images/average_median_acc.pdf}
\includegraphics[width=.32\textwidth]{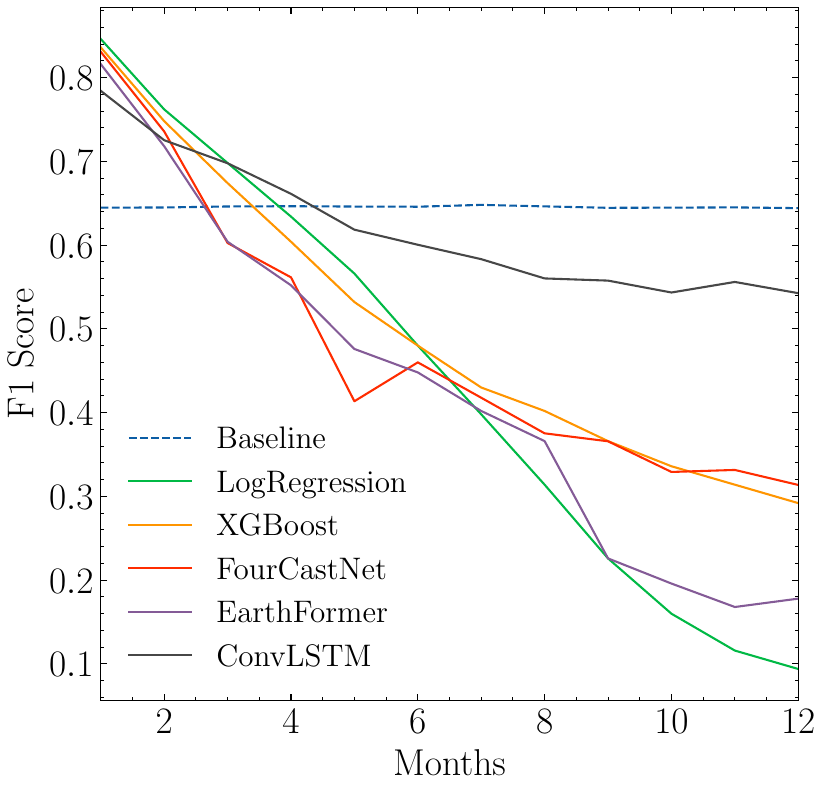}
    \caption{Quality metrics for the binary drought severity classification: median $\mathrm{ROC\, AUC}$, $\mathrm{PR\, AUC}$, and $\mathrm{F1}$  for different forecast horizons averaged over five considered regions}
    \label{fig:metrics_comparison_binary}
\end{figure}

\paragraph{Analysis of results}
\label{sec:analysis_main_results}
The primary results are depicted in Figure~\ref{fig:metrics_comparison_binary}. 
Our findings indicate that EarthFormer outperforms other approaches for shorter horizons. In particular, EarthFormer reached $ROC\,AUC$ score of $0.95$ for a one-month prediction. But EarthFormer falls short of ConvLSTM at longer horizons of 9-12 months. The ConvLSTM model showed the second-best result (after EarthFormer), achieving an impressive $ROC\,AUC$ score of $0.9$ for a one-month prediction. Notably, ConvLSTM exhibits a gradual decline in performance, reaching $0.6-0.65$ for longer forecasting horizons (ranging from 9 to 12 months) but nevertheless beating all other models. The standard gradient boosting approach initially yields a similar $ROC\, AUC$ score of $0.9$ but sharply drops to $0.5$ as the forecasting horizon is extended. 

%Accuracy scores reflect better performance of logistic regression and XGBoost, but we shouldn't rely on these results as this metric does not indicate unbalanced class prediction. 

Additionally, we present the results for six-month predictions by regions in Figure~\ref{tab:medianmetrics_region_binary_6m}, where we show the variation in scores for different geographies across models.

\textit{Why does transformer fail in long-term prediction?} Our assumption for such behavior is the permutation-invariance of the attention mechanism. Despite positional encoding, transformers cannot effectively extract temporal information from long input sequences. Since a long input sequence is essential for long-term forecasting, transformers do not show good results. Similar results for different time-series tasks were observed in ~\cite{zeng2023linear}. ConvLSTM naturally extracts temporal information via LSTM.

\textit{Why is logistic regression impressive?} Since gradient boosting results are almost identical to those of logistic regression, we discuss only logistic regression performance. First, the power of linear models was already shown in ~\cite{zeng2023linear}, where they beat modern transformer architectures on almost all datasets. In our experiments, linear models are worse than other models on \textit{long-term} prediction, but on a \textit{short-term} scale, we can see comparable results. We tried different history lengths, but our results show that taking only the element nearest to the future horizons is sufficient. Our intuition is that the nearest future predictor variables are closely (particularly linearly) related to the history element. For example, PDSI in July is close to PDSI in August but far away from December. Hence, linear models are good at \textit{short-term} predictions but poor at \textit{long-term} forecasting.

\begin{table}[!h]
\small
    \centering
    \begin{tabular}{rccccc}
         \hline
         Horizon, months & 1 & 3 & 6 & 9 & 12 \\
         \hline
         Median $\mathrm{ROC\, AUC}$: &  &   &   &   &   \\
         \hline
         Baseline & 0.5 & 0.5 & 0.5 & 0.5 & 0.5 \\
         LogRegression & 0.886 & 0.774  & 0.640  & 0.546  & 0.518 \\
         XGBoost & 0.878 & 0.754 & 0.628 & \underline{0.568} & \underline{0.542} \\
         FourCastNet & 0.881  & 0.711 & 0.624 & 0.561 & 0.536 \\
         EarthFormer & \textbf{0.948} & \textbf{0.840} & \underline{0.690} & 0.556 & 0.480 \\
         ConvLSTM & \underline{0.887} & \underline{0.802} & \textbf{0.693} & \textbf{0.650} & \textbf{0.617} \\
         \hline
         Median PR AUC: &  &   &   &   &   \\
         \hline 
         Baseline & 0.355 & 0.354 & 0.354 & 0.355 & 0.356 \\
         LogRegression & 0.766  & 0.598  & 0.470  & 0.394  & 0.360 \\
         XGBoost & 0.752 & 0.574 & 0.44 & 0.39 & 0.372 \\
         FourCastNet & \underline{0.776} & 0.546 & 0.455 & 0.402 & \underline{0.382} \\
         EarthFormer & \textbf{0.880} & \underline{0.650} & \underline{0.514} & \underline{0.438} & 0.362 \\
         ConvLSTM & 0.772 & \textbf{0.689} & \textbf{0.565} & \textbf{0.505} & \textbf{0.452} \\
         \hline
         Median F1: &  &   &   &   &   \\
         \hline
         Baseline & 0.645 & 0.646 & \textbf{0.646} & \textbf{0.645} & \textbf{0.644} \\
         LogRegression & \textbf{0.846} & \textbf{0.698}  & 0.480  & 0.226  & 0.094 \\
         XGBoost & \underline{0.836} & \underline{0.674} & 0.480 & 0.366 & 0.292 \\
         FourCastNet & 0.831 & 0.603 & 0.460 & 0.366 & 0.314 \\
         EarthFormer & 0.816 & 0.604 & 0.448 & 0.226 & 0.178 \\
         ConvLSTM & 0.784 & \textbf{0.698} & \underline{0.600} & \underline{0.558} & \underline{0.543} \\
         \hline
         % Median Accuracy: &  &   &   &   &   \\
         % \hline
         % Baseline & 0.640 & 0.640 & 0.639 & 0.639 & 0.640 \\
         % LogReg & \underline{0.908} & 0.814 & \underline{0.730} & \underline{0.684} & \underline{0.676} \\
         % XGBoost & \textbf{0.910} & \textbf{0.824} & \textbf{0.736} & \textbf{0.692} & \textbf{0.682} \\
         % FourCastNet & 0.835 & 0.703 & 0.616 & 0.583 & 0.547 \\
         % EarthFormer & 0.906 & \underline{0.818} & 0.714 & 0.616 & 0.578 \\
         % ConvLSTM & 0.816 & 0.755 & 0.672 & 0.619  & 0.599 \\
         % \hline
    \end{tabular}
    \caption{Median Metrics vs. Forecast Horizon, binary classification; best values are in bold, second best are underlined}
    \label{tab:medianmetrics_binary}
\end{table}

\begin{table}[!h]
\small
    \centering
    \begin{tabular}{rccccc}
        \hline
         Region & Northern & Poland & Madhya & Goias & Missouri \\
          & Kazakhstan &  & Pradesh & & \\
         \hline
         Median $\mathrm{ROC\, AUC}$: &  &   &   &   &   \\
         \hline
         Baseline & 0.5 & 0.5 & 0.5 & 0.5 & 0.5 \\
         LogReg & \underline{0.60} & 0.63 & 0.61 & \underline{0.61}  & \textbf{0.75} \\
         XGBoost & 0.59 & 0.64 & 0.59 & \underline{0.61}  & 0.71 \\
         FourCastNet & \underline{0.60} & 0.55 & 0.65 & \underline{0.61} & 0.69\\
         EarthFormer & 0.54 & \textbf{0.69} & \textbf{0.84} & \textbf{0.65} & \underline{0.73} \\
         ConvLSTM &  \textbf{0.71} & \underline{0.68} & \underline{0.71}  &  0.60 & \textbf{0.75} \\
         \hline
         Median PR AUC: &  &   &   &   &   \\
         \hline
         Baseline & 0.37  & 0.43  & 0.35 & 0.39 & 0.23 \\
         LogReg & 0.46 & \textbf{0.67} & 0.24 & \textbf{0.55}   & 0.43 \\
         XGBoost & 0.46 & 0.63 & 0.20 & \underline{0.54}  & 0.37 \\
         FourCastNet & 0.46 & 0.55 & 0.36  & \underline{0.54}  & 0.37 \\
         EarthFormer & \underline{0.47} & 0.22 & \textbf{0.77} & 0.52 & \textbf{0.59}\\
         ConvLSTM & \textbf{0.55} & \underline{0.65} & \underline{0.57} & \underline{0.54} & \underline{0.50} \\
         \hline
         Median F1: &  &   &   &   &   \\
         \hline
         Baseline & \underline{0.63} & 0.57 & \underline{0.65} & \textbf{0.61} & \textbf{0.77} \\
         LogReg & 0.42 & \underline{0.62} & 0.35 & 0.41 & 0.60 \\
         XGBoost & 0.45 & 0.58 & 0.30 & 0.54 & 0.53 \\
         FourCastNet & 0.42 & 0.39 & 0.46 & 0.51 & 0.52 \\
         EarthFormer & 0.03 & 0.18 & \textbf{0.80} & \underline{0.55} & \underline{0.65} \\
         ConvLSTM & \textbf{0.65} & \textbf{0.64} & 0.64 & 0.52  & 0.56  \\
         \hline
    \end{tabular}
    \caption{Median Metrics vs. Region, binary classification, six-month horizon; best values are in bold, second best are underlined}
    \label{tab:medianmetrics_region_binary_6m}
\end{table}

\subsection{Predictions and errors for a particular region}

\begin{figure}[!h]
    \centering
    \includegraphics[width=0.5\textwidth]{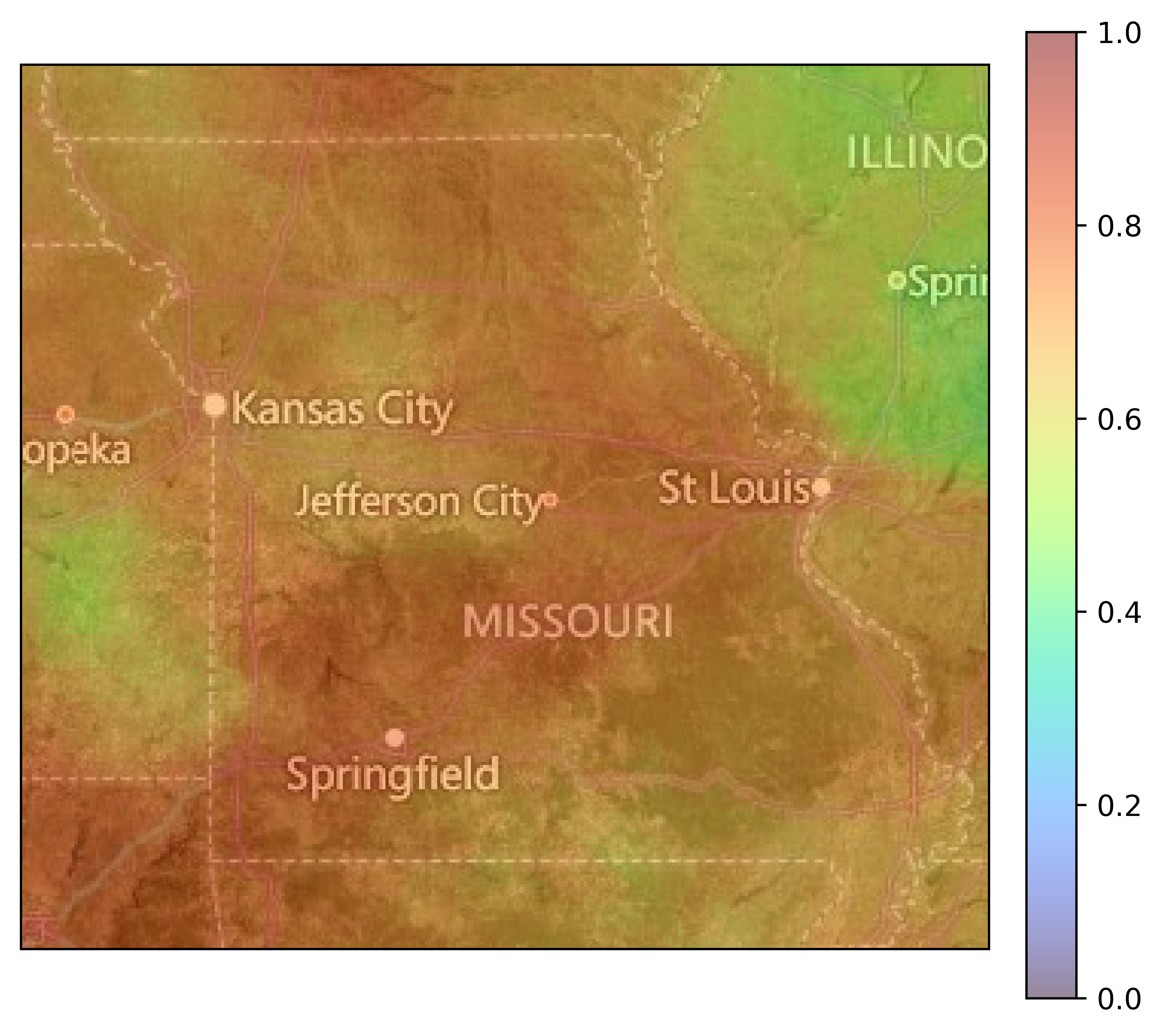}
    \caption{Spatial distribution of $\mathrm{ROC\, AUC}$ for 6 month forecast, Missouri, ConvLSTM}
    \label{fig:rocauc_spatial}
\end{figure}

To assess the performance of the Convolutional LSTM algorithm (which proved to be the most stable and promising for long-term drought forecasting), we focused on the region of Missouri, where we ran several ablation studies. To illustrate, the spatial distribution of $ROC\, AUC$ scores is depicted in Figure~\ref{fig:rocauc_spatial}. Notably, we observed a non-uniform distribution of $ROC\, AUC$ values across the cells within the region. The standard deviation of the scores is substantial, and individual values range from close to random predictors ($ROC\, AUC=0.6$) to near-perfect scores approaching $1.0$. This variability highlights the diverse predictive capabilities of our algorithm across different spatial locations within Missouri.

\subsubsection{Performance Evaluation for Cropped Region}
\paragraph{Description of experiment}
As is typical with $ROC\, AUC$  maps, the worst predictions are found on the edges and some corners. We have observed that this behavior is consistent regardless of the region being studied. Consequently, making predictions for a larger region and cropping the desired region of interest may be advantageous. We have conducted a study to test this hypothesis for Figure~\ref{fig:rocauc_spatial}, and the results are shown in Table~\ref{tab:crop}.

\begin{table}[!h]
\small
    \centering
    \begin{tabular}{lccccc}
         \hline
         Percent of map cropped & 0 & 10 & 20 & 30 & 40 \\
         
         median $\mathrm{ROC\, AUC}$ & 0.7525 & 0.7592 & 0.7665 & 0.7749 & 0.7834  \\
         \hline
         Percent of map cropped & 50 & 60 & 70 & 80 & 90 \\
         
         median $\mathrm{ROC\, AUC}$ & 0.7886 & 0.7899 & 0.7880 & 0.7838 & 0.7825 \\
         \hline
    \end{tabular}
    \caption{$\mathrm{ROC\, AUC}$ score vs. crop percentage, six-month forecast, ConvLSTM model for Missouri}
    \label{tab:crop}
\end{table}

\paragraph{Analysis of results}
Based on this experiment's findings, we deduce that cropping approximately 40-50\% of the initially selected region maximizes our score. In other words, choosing a region that is initially 1.6-2 times larger than our target region is advisable. However, the precise amount of zoom required for optimal results must be determined through further experiments.\\
Next, we conducted several similar experiments to investigate how the model predictions change with the decrease in the total squared area. We took the same geographic region, Missouri, and examined various combinations of history length and forecast horizon. We trained a new model for each history length variant, forecast horizon, and region area (varying from the entire state to about a quarter of it).

\begin{figure*}
    \includegraphics[width=.25\textwidth]{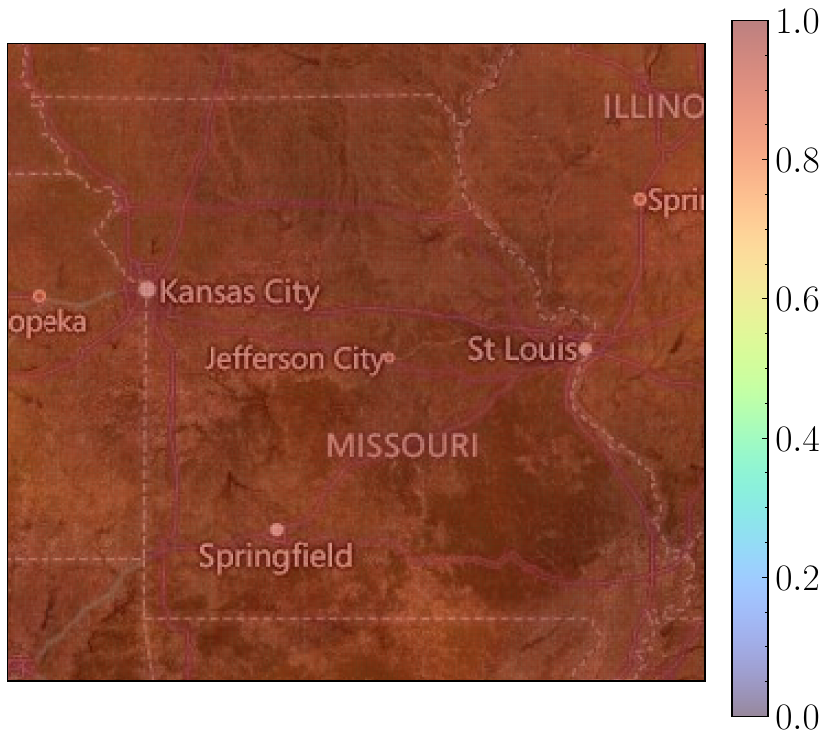}\hfill
    \includegraphics[width=.25\textwidth]{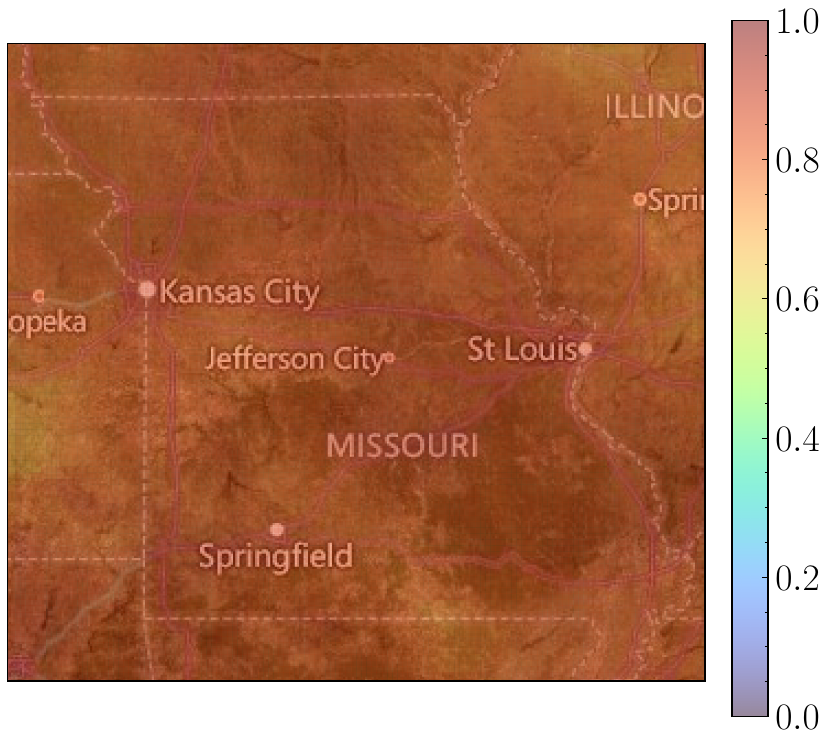}\hfill
    \includegraphics[width=.25\textwidth]{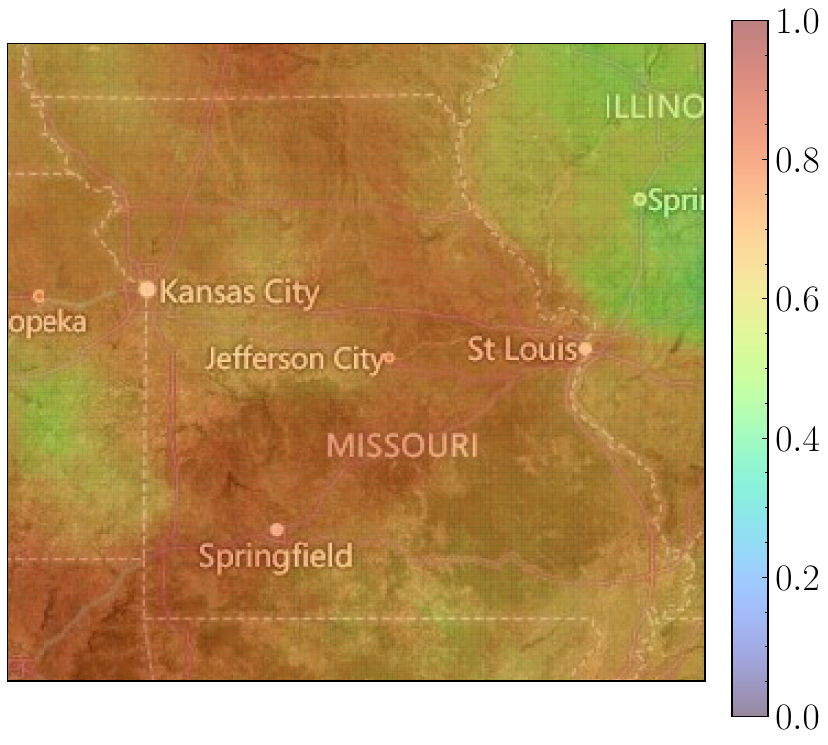}\hfill
    \includegraphics[width=.25\textwidth]{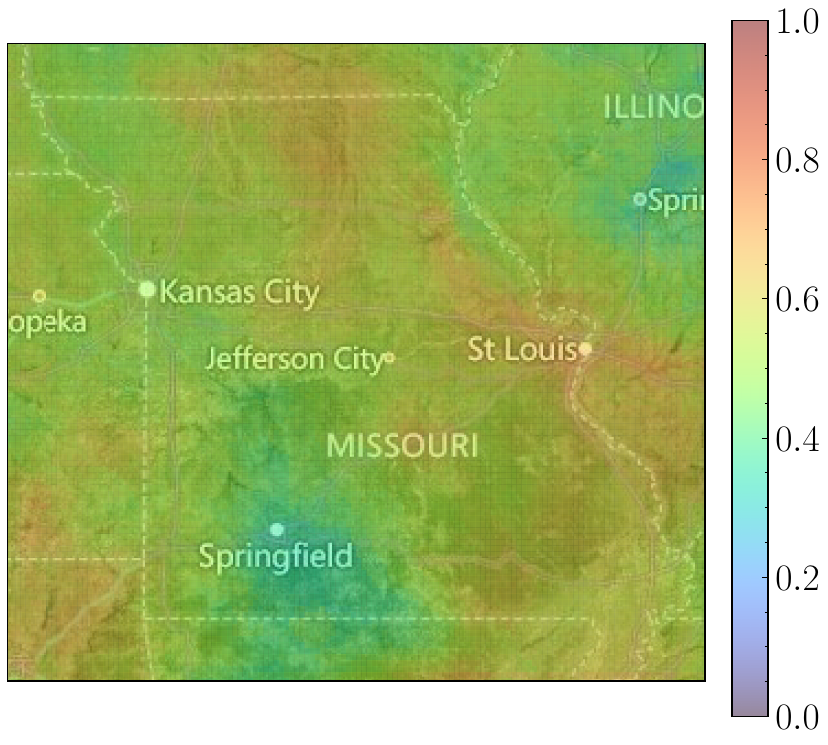}
    \\[\smallskipamount]
    \includegraphics[width=.25\textwidth]{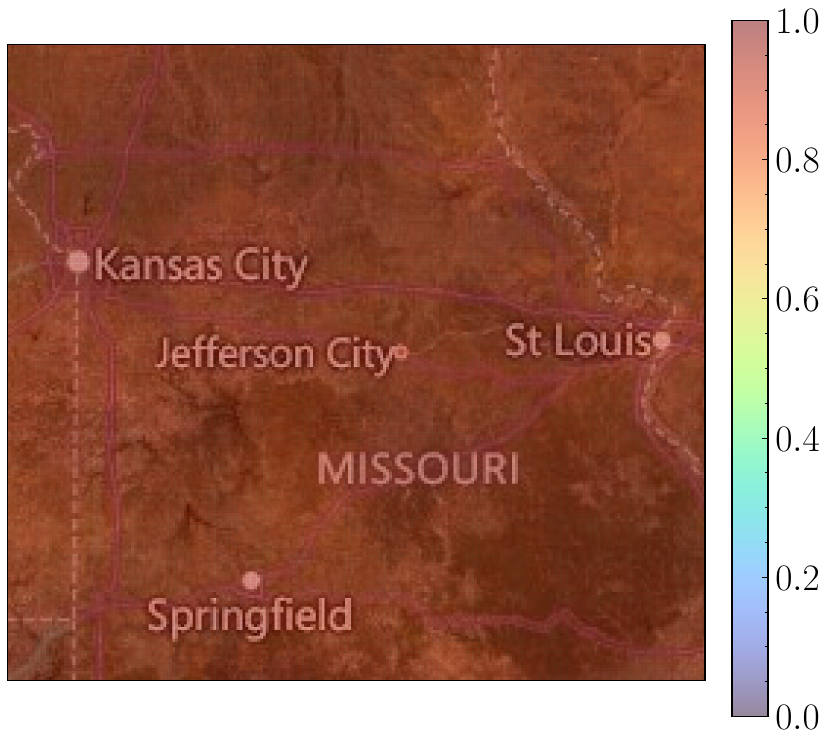}\hfill
    \includegraphics[width=.25\textwidth]{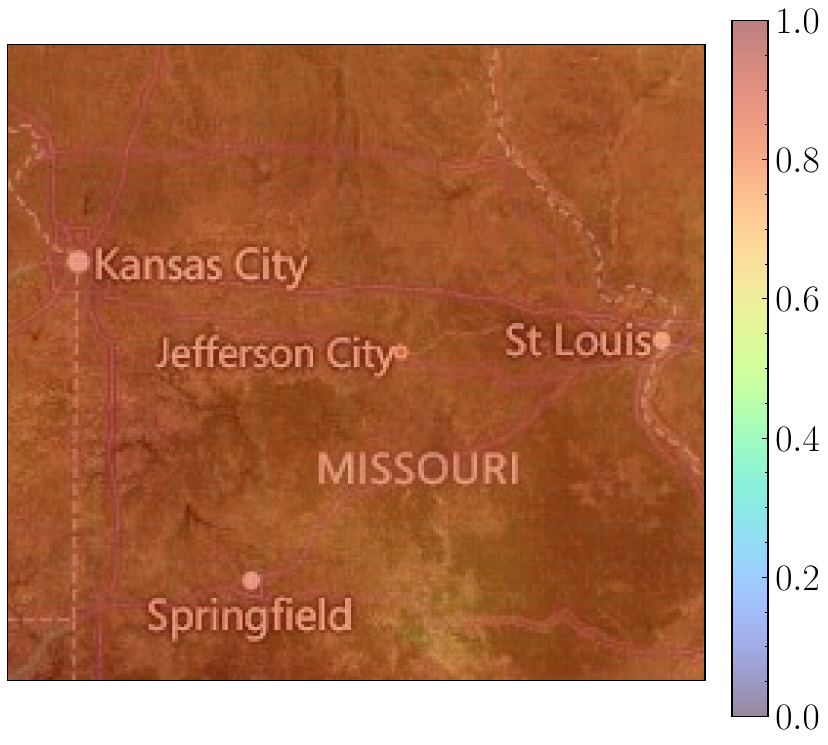}\hfill
    \includegraphics[width=.25\textwidth]{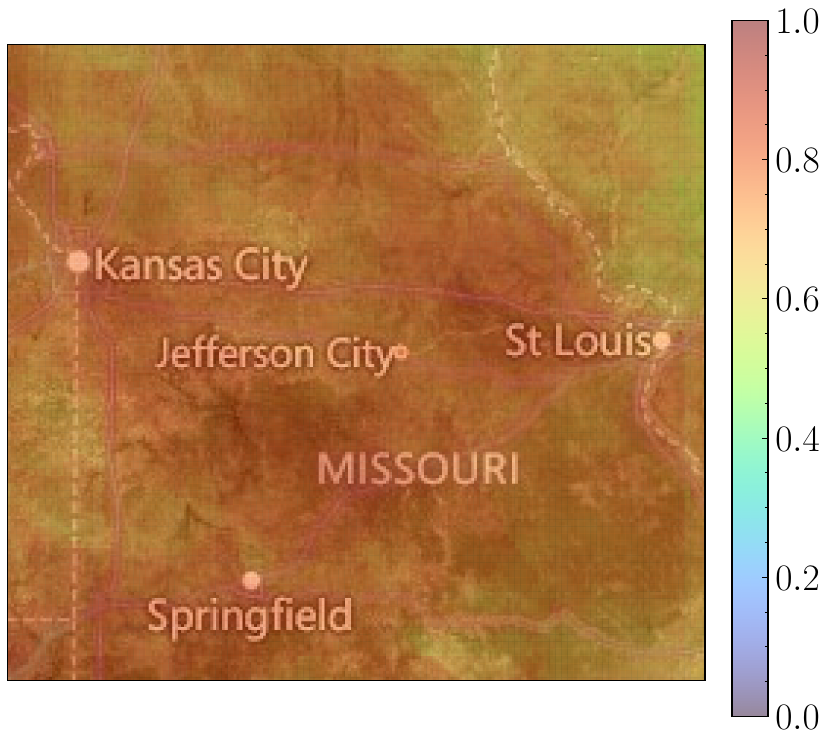}\hfill
    \includegraphics[width=.25\textwidth]{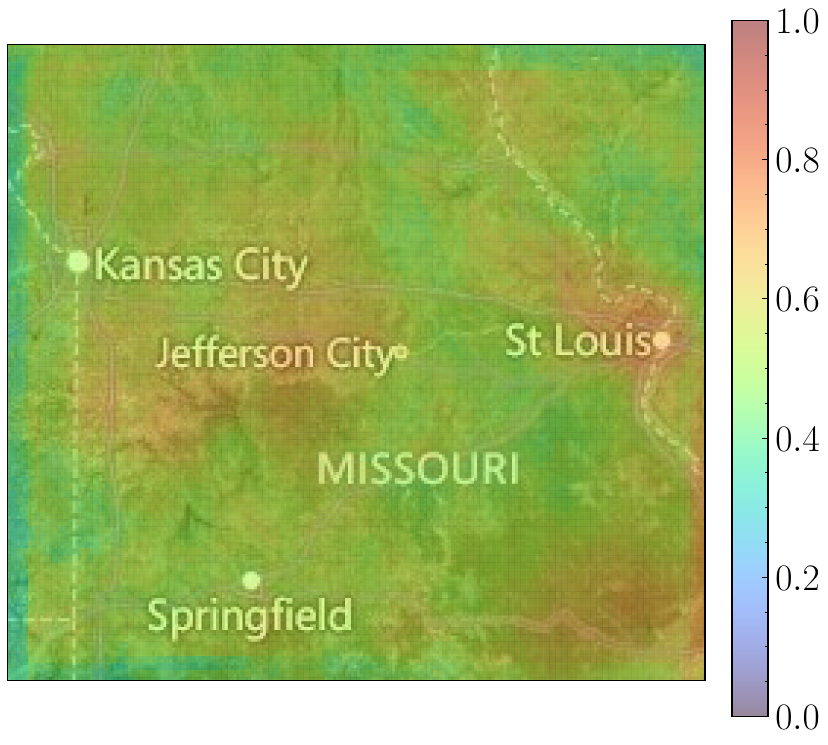}
    \\[\smallskipamount]
    \includegraphics[width=.25\textwidth]{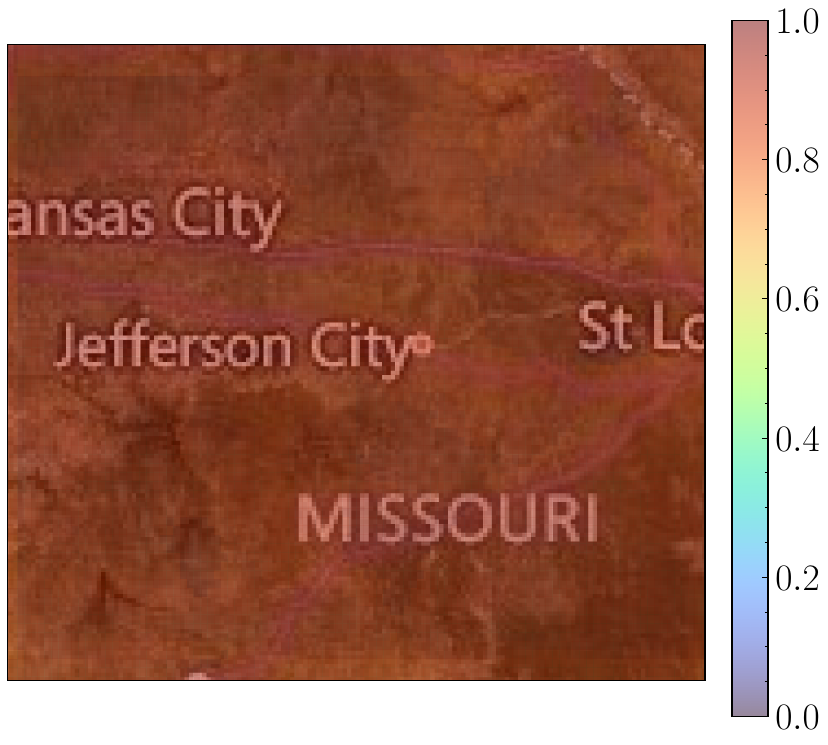}\hfill
    \includegraphics[width=.25\textwidth]{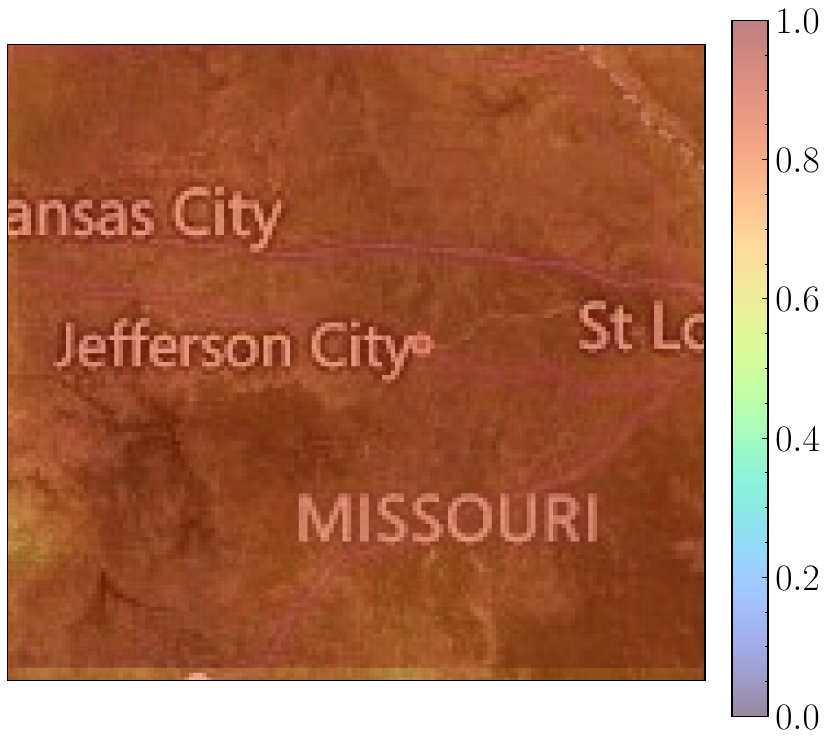}\hfill
    \includegraphics[width=.25\textwidth]{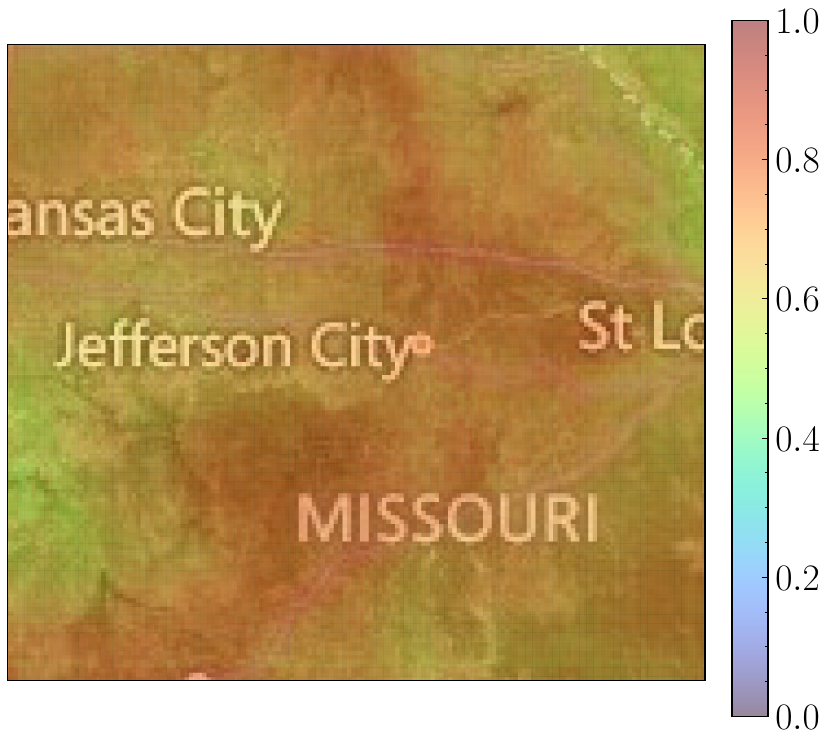}\hfill
    \includegraphics[width=.25\textwidth]{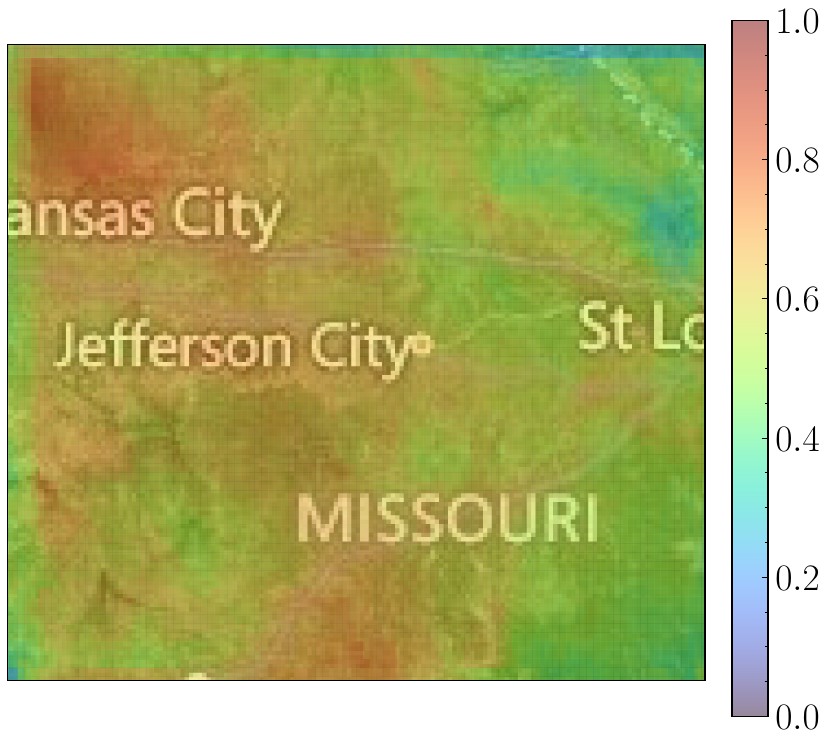}
    \\[\smallskipamount]
    \includegraphics[width=.25\textwidth]{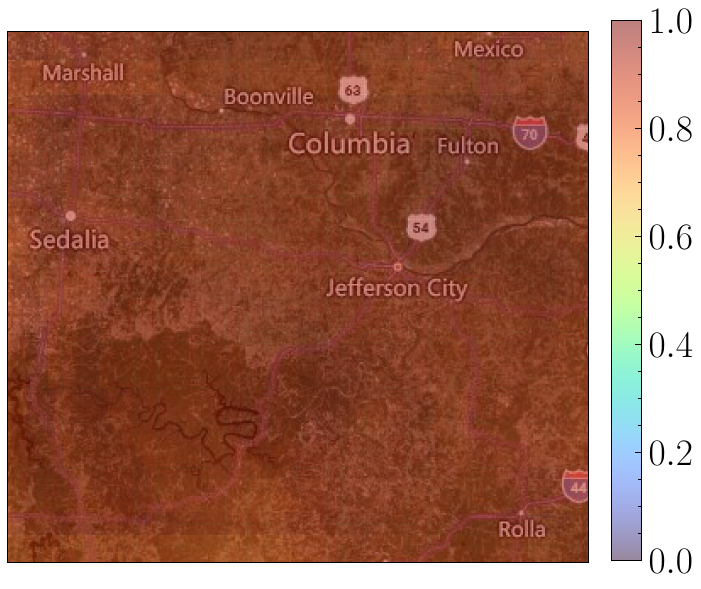}\hfill
    \includegraphics[width=.25\textwidth]{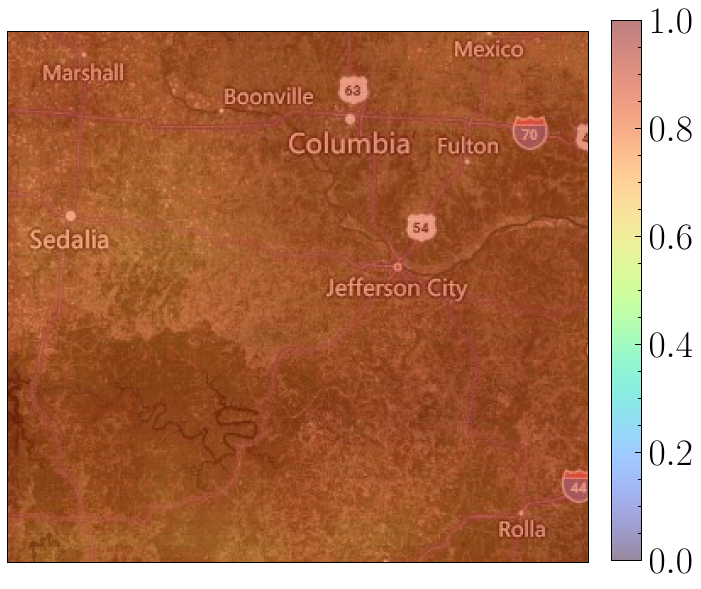}\hfill
    \includegraphics[width=.25\textwidth]{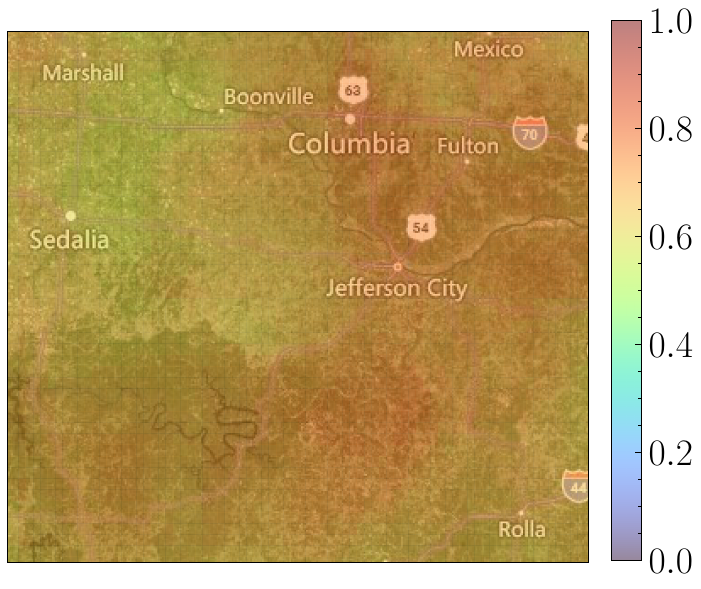}\hfill
    \includegraphics[width=.25\textwidth]{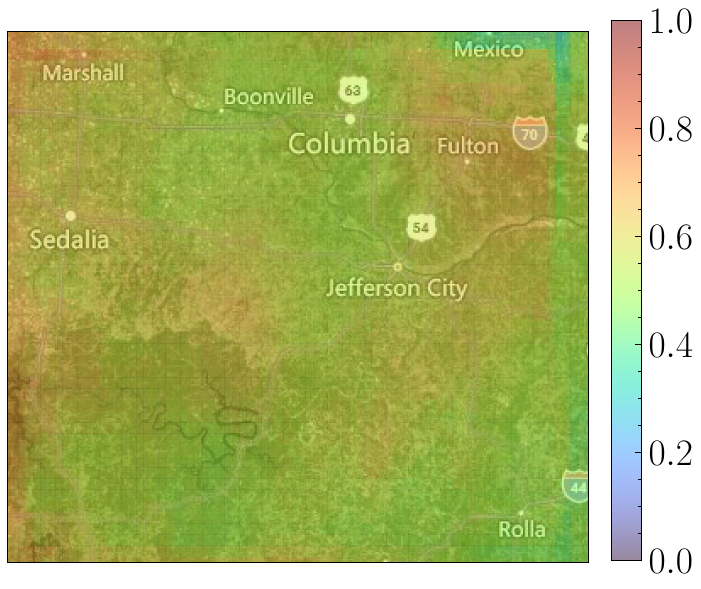}
    \caption{Evolution of spatial maps with zooming in: top to bottom area decreases from 100\% of the region to 27\%, left to right forecast horizons are 1, 3, 6, and 12 months correspondingly.}\label{fig:zoom_2}
\end{figure*}

The results are summarized in Table~\ref{tab:zoom_in}, and they are presented in more detail in Tables~\ref{tab:zoom_in_6x1},~\ref{tab:zoom_in_12x3},~\ref{tab:zoom_in_9x6}, and ~\ref{tab:zoom_in_24x12}. We observed that predicting a smaller area with a pretrained model from a larger area generally works better. However, the degree of improvement is marginal, usually not exceeding 0.5-1\%. Figure~\ref{fig:zoom_2} presents the evolution of spatial maps.

\begin{table}[!h]
\small
    \centering
    \begin{tabular}{lccccc}
         \hline
         Region area & 100\% & 75\% & 53\% & 27\% \\
         \hline
         $h = 6$, $f = 1$ & 0.9292 & 0.9466 & 0.9432 & 0.9326  \\
         $h = 12$, $f = 3$ & 0.8746 & 0.8525  & 0.8595 & 0.8710 \\
         $h = 9$, $f = 6$ & 0.7525 & 0.6901 & 0.7597  & 0.7544 \\
         $h = 24$, $f = 12$ & 0.6117 & 0.5890 & 0.6126 & 0.6264  \\
         \hline
    \end{tabular}
    \caption{$\mathrm{ROC\, AUC}$ score vs zoomed region area (h - length of input history, f - forecast horizon)}
    \label{tab:zoom_in}
\end{table}

\subsubsection{Standard deviations and ensembling}
\paragraph{Description of experiment}
To assess the consistency of our findings, we repeated the experiment of PDSI binary classification for Missouri with five random seeds for each configuration of history length and forecast horizon. In addition, we tested an averaged ensemble of these five trained models.

\begin{figure*}
     \centering
     \includegraphics[width=\textwidth]{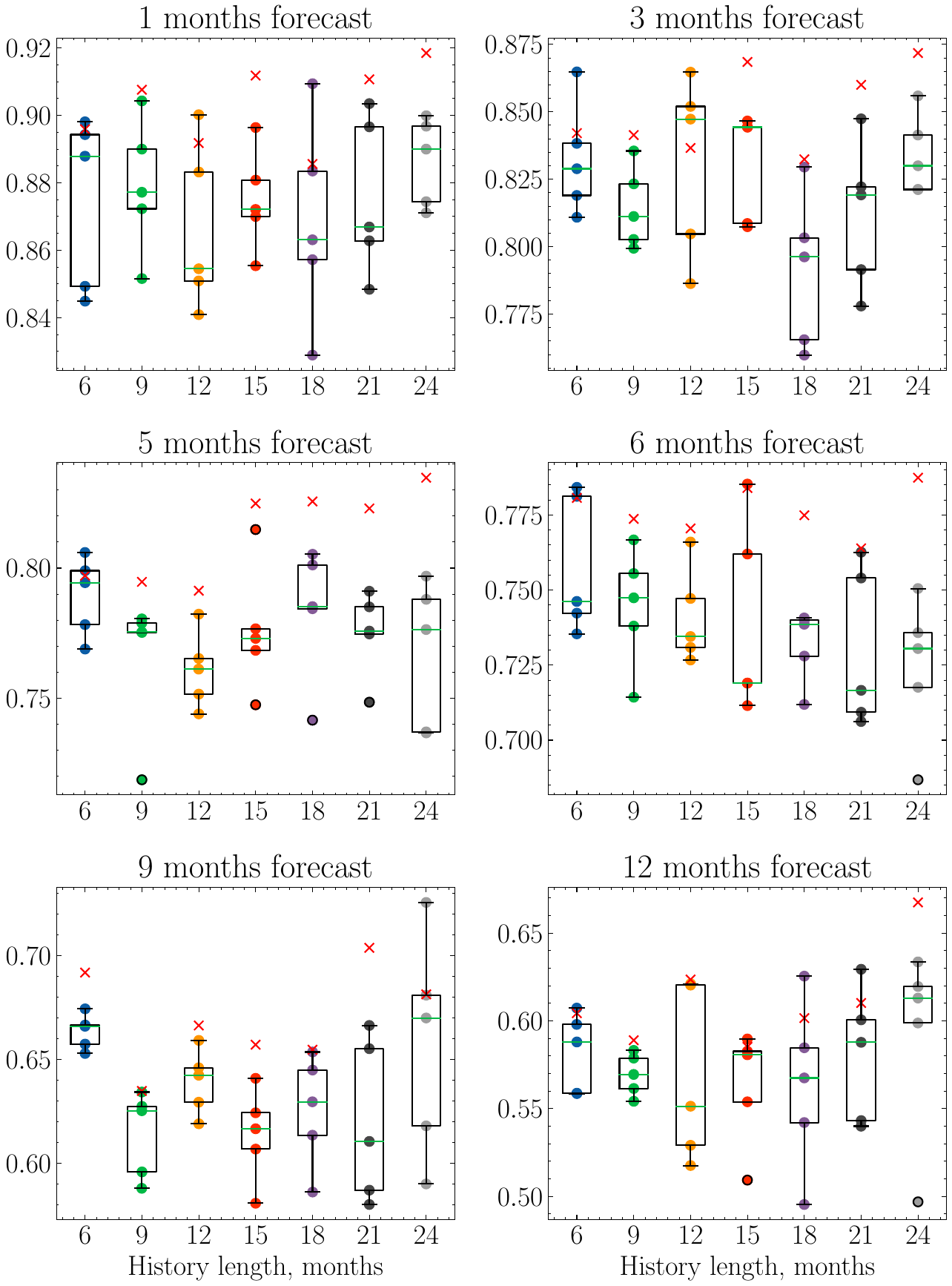}
    \caption{Variation of $\mathrm{ROC\, AUC}$ scores for five different random seeds; ConvLSTM binary classification for Missouri state. Metrics for an ensemble of five models are marked as red crosses.}
    \label{fig:seeds}
\end{figure*}

\paragraph{Analysis of results}
The results for horizons of 1, 3, 5, 6, 9, and 12 months can be found in Figure~\ref{fig:seeds}. Ensemble scores are denoted by the red crosses on these plots. Overall, the numbers vary, and there is no definitive optimal choice of history length for any horizon. However, we observed that extreme values yield better performance, such as the shortest or longest history lengths (e.g., 6, 9, 21, and 24 months). Notably, the averaged ensemble of models outperforms the individual underlying models in most cases.
We note that this effect is more sound for neural networks, as they provide more diverse predictions, even if there are no differences in the architecture, and the only difference is a starting point for training~\citep{fort2019deep}.
On the contrary, the logistic regression ensemble is similar to a single model, as the optimization problem is convex~\citep{bishop2006pattern}, and gradient boosting is an ensemble per se~\citep{mehr2021xgboost}.
The obtained ensembles can also be used to access the uncertainty of predictions by machine learning models, improving the decision-making process~\citep{jain2020maximizing}.

% \subsection{Usage of a climate model for fusing and better understanding}
% ?

\section{Conclusion}
Droughts, increasingly severe and frequent due to climate change, pose significant threats to agriculture and public health. The summer of 2022 in the Northern Hemisphere highlighted the urgency of these issues. Our research focuses on improving drought forecasting, a crucial step in mitigating the adverse impacts of these natural disasters. To tackle the task, we employed various models (from classic models to modern transformers) and many distinct regions to test their performance.

We succeeded in providing a better model suitable for agricultural decision-making and insurance applications.
Our variant of EarthFormer shows the best result in \textit{short-term} forecasting. In one-month prediction $ROC\, AUC$ score is $0.948$.
Our variant of ConvLSTM is much better than other models in \textit{long-term} forecasting, achieving an impressive $ROC\, AUC$ score of $0.617$ in twelve-month prediction.
The metric values above are much higher than classic approaches: we significantly reduced the gap between perfect $ROC\, AUC$ score and ours on $54\%$ and $16\%$, for short and long-term predictions, respectively. 
Before our study, $12$-months ahead prediction gave close to random results, which is no longer the case.
Additional improvement can be obtained by using an ensemble of deep learning models and increasing the amount of used data.

So we recommend to use EarthFormer for \textit{short-term} predictions and ConvLSTM for \textit{long-term}.
For better predictions, one should use an ensemble for such models.
Such a combination leads to a good model for the considered time horizons.

\section*{Acknowledgements}

 The work of A. Marusov, V.Grabar, A.Zaytsev and A.Bulkin was supported by the Analytical Center for the Government of Russian Federation (subsidy agreement 000000D730321P5Q0002, Grant No. 70-2021-00145 02.11.2021).

% \appendix

% \section{Sample Appendix Section}
% \label{sec:sample:appendix}
% Lorem ipsum dolor sit amet, consectetur adipiscing elit, sed do eiusmod tempor section \ref{sec:sample1} incididunt ut labore et dolore magna aliqua. Ut enim ad minim veniam, quis nostrud exercitation ullamco laboris nisi ut aliquip ex ea commodo consequat. Duis aute irure dolor in reprehenderit in voluptate velit esse cillum dolore eu fugiat nulla pariatur. Excepteur sint occaecat cupidatat non proident, sunt in culpa qui officia deserunt mollit anim id est laborum.

 \bibliography{cas-refs}




\section{Appendix}

\subsection{Multiclass problem}
\paragraph{Description of experiment}
As an additional experiment and study of models' limits, we looked at the multiclass classification problem for Missouri. For the 3 class problem, we arbitrarily set up thresholds of PDSI values as $-1,\,\,1$ and for the 5 class as $-3,\,\,-1,\,\,1,\,\,3$. As an evaluation metric, we use median accuracy over all celled predictions.
For gradient boosting and logistic regression, we use default implementations for multiclass predictions. For the neural networks-based model, we replace two possible output cells with the number of cells equal to the number of classes.

\begin{figure}[!h]
    \centering
\includegraphics[width=.50\textwidth]{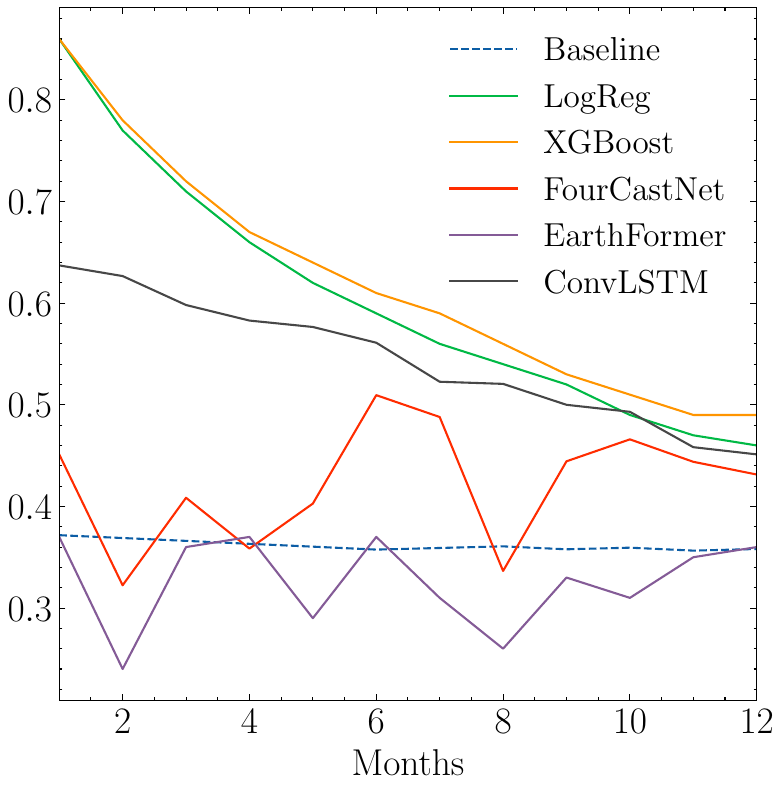}\hfill
\includegraphics[width=.50\textwidth]{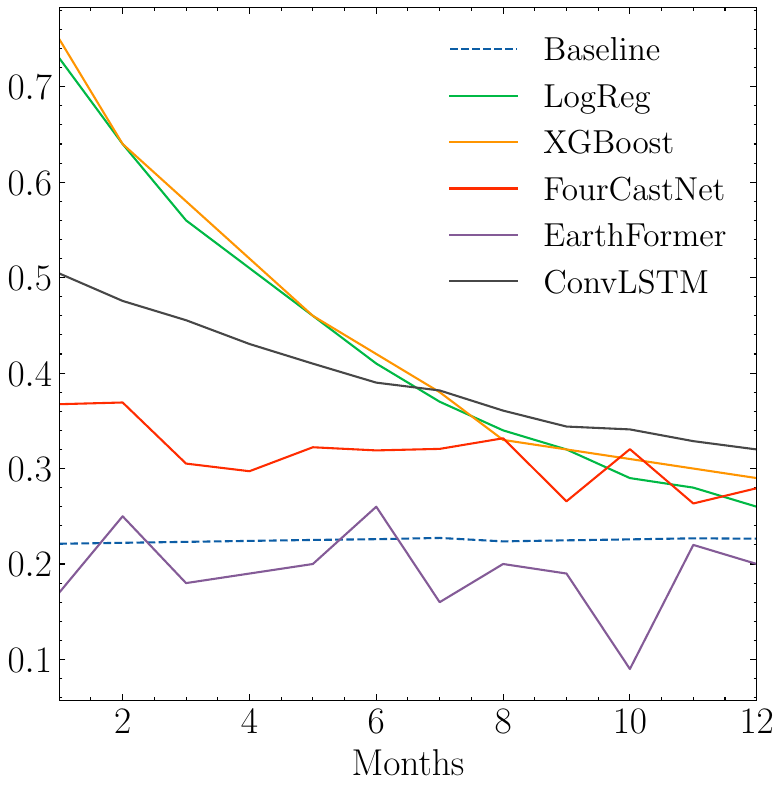}
    \caption{Accuracy vs. forecast horizon, 3 and 5 class task, Missouri}
    \label{fig:metrics_comparison_multi}
\end{figure}

\paragraph{Analysis of results}
Results, provided in Figure~\ref{fig:metrics_comparison_multi} and Table~\ref{tab:medianmetrics_multi}, are similar to a binary problem. Logistic regression and gradient boosting hold their superiority longer, the Convolutional LSTM score is relatively stable, and only the transformers' prediction disappoints, falling to the level of the historical baseline.

\begin{table}[!h]
\small
    \centering
    \begin{tabular}{rccccc}
        \hline
         Horizon, months & 1 & 3 & 6 & 9 & 12 \\
         \hline
         median Accuracy, 3 class: &  &   &   &   &   \\
         \hline
         Baseline & 0.372 & 0.366 &  0.358 & 0.358 &  0.358 \\
         LogReg & \textbf{0.86} & \underline{0.71} & \underline{0.59} & \underline{0.52} & \underline{0.46} \\
         XGBoost & \textbf{0.86} & \textbf{0.72} & \textbf{0.61} & \textbf{0.53} & \textbf{0.49} \\
         FourCastNet & 0.451 & 0.409  & 0.509  & 0.444  & 0.431 \\
         EarthFormer & 0.37 & 0.36 & 0.37 &  0.33 & 0.36 \\
         ConvLSTM & \underline{0.637} & 0.598 & 0.561 & 0.500 & 0.451 \\
         \hline
         median Accuracy, 5 class: &  &   &   &   &   \\
         \hline
         Baseline & 0.221 & 0.223  &  0.226 & 0.225 & 0.223  \\
         LogReg & \underline{0.73} & \underline{0.56} & \underline{0.41} & \underline{0.32} & 0.26 \\
         XGBoost & \textbf{0.75} & \textbf{0.58} & \textbf{0.42} & \underline{0.32} & \underline{0.29}  \\
         FourCastNet & 0.367  & 0.305  & 0.319  &  0.266 & 0.279 \\
         EarthFormer & 0.17 & 0.18 & 0.26 & 0.19 &  0.20 \\
         ConvLSTM & 0.504 & 0.455  & 0.389 & \textbf{0.344} &  \textbf{0.312} \\
         \hline
    \end{tabular}
    \caption{Median Accuracy vs. Forecast Horizon, 3 and 5 possible class problems, Missouri; best values are in bold, second best are underlined}
    \label{tab:medianmetrics_multi}
\end{table}

\begin{table}[!h]
\small
    \centering
    \begin{tabular}{lcccc}
         \hline
         Region & 104x136 & 88x120 & 72x104 & 48x80  \\
         area & (100\%) & (75\%) & (53\%) & (27\%)  \\
         \hline
         104x136 (100\%) & 0.9292 & 0.9302  & 0.9327  & 0.9356    \\
         88x120 (75\%) & - & 0.9466 & 0.9487 & 0.9515    \\
         72x104 (53\%) & - & - & 0.9432 & 0.9438  \\
         48x80 (27\%) & - & - & - & 0.9326   \\
         \hline
    \end{tabular}
    \caption{$\mathrm{ROC\, AUC}$ score trained on a subset (left) and evaluated on a different region (top) (length of input history = 6, forecast horizon = 1)}
    \label{tab:zoom_in_6x1}
\end{table}

\begin{table}[!h]
\small
    \centering
    \begin{tabular}{lcccc}
         \hline
         Region & 104x136 & 88x120 & 72x104 & 48x80  \\
         area & (100\%) & (75\%) & (53\%) & (27\%)  \\
         \hline
         104x136 (100\%) & 0.8746  & 0.8775  & 0.8816  & 0.8829   \\
         88x120 (75\%) & - & 0.8525  & 0.8536 & 0.8508    \\
         72x104 (53\%) & - & - & 0.8595 & 0.8655   \\
         48x80 (27\%) & - & - & - &  0.8710  \\
         \hline
    \end{tabular}
    \caption{$\mathrm{ROC\, AUC}$ score trained on a subset (left) and evaluated on a different region (top) (length of input history = 12, forecast horizon = 3)}
    \label{tab:zoom_in_12x3}
\end{table}

\begin{table}[!h]
\small
    \centering
    \begin{tabular}{lcccc}
         \hline
         Region & 104x136 & 88x120 & 72x104 & 48x80  \\
         area & (100\%) & (75\%) & (53\%) & (27\%)  \\
         \hline
         104x136 (100\%) & 0.7525  & 0.7624  &  0.7722 & 0.7897   \\
         88x120 (75\%) & - & 0.6901  & 0.6974 &  0.7081   \\
         72x104 (53\%) & - & - & 0.7597  & 0.7794    \\
         48x80 (27\%) & - & - & - & 0.7544  \\
         \hline
    \end{tabular}
    \caption{$\mathrm{ROC\, AUC}$ score trained on a subset (left) and evaluated on a different region (top) (length of input history = 9, forecast horizon = 6)}
    \label{tab:zoom_in_9x6}
\end{table}

\begin{table}[!h]
\small
    \centering
    \begin{tabular}{lcccc}
         \hline
         Region & 104x136 & 88x120 & 72x104 & 48x80  \\
         area & (100\%) & (75\%) & (53\%) & (27\%)  \\
         \hline
         104x136 (100\%) & 0.6117 & 0.6154 & 0.6142  & 0.6210    \\
         88x120 (75\%) & - & 0.5890  & 0.5980 & 0.6033    \\
         72x104 (53\%) & - & - & 0.6126  & 0.6184    \\
         48x80 (27\%) & - & - & - & 0.6264    \\
         \hline
    \end{tabular}
    \caption{$\mathrm{ROC\, AUC}$ score trained on a subset (left) and evaluated on a different region (top) (length of input history = 24, forecast horizon = 12)}
    \label{tab:zoom_in_24x12}
\end{table}

% \end{thebibliography}
\end{document}